# Neuromimetic metaplasticity for adaptive continual learning

Suhee Cho[1], Hyeonsu Lee[2], Seungdae Baek[2] & Se-Bum Paik[1,2*]

[1]Department of Brain and Cognitive Sciences, [2]Department of Bio and Brain Engineering, Korea Advanced Institute of Science and Technology, Daejeon 34141, Republic of Korea

*Correspondence: Se-Bum Paik (sbpaik@kaist.ac.kr)

## Abstract

Conventional intelligent systems based on deep neural network (DNN) models encounter challenges in achieving human-like continual learning due to catastrophic forgetting. Here, we propose a metaplasticity model inspired by human working memory, enabling DNNs to perform catastrophic forgetting-free continual learning without any pre- or post-processing. A key aspect of our approach involves implementing distinct types of synapses from stable to flexible, and randomly intermixing them to train synaptic connections with different degrees of flexibility. This strategy allowed the network to successfully learn a continuous stream of information, even under unexpected changes in input length. The model achieved a balanced tradeoff between memory capacity and performance without requiring additional training or structural modifications, dynamically allocating memory resources to retain both old and new information. Furthermore, the model demonstrated robustness against data poisoning attacks by selectively filtering out erroneous memories, leveraging the Hebb repetition effect to reinforce the retention of significant data.

# 1   Introduction

Recent advances in deep neural network (DNN) models have demonstrated remarkable performance outcomes across a wide range of tasks, often surpassing human capabilities in areas such as image recognition [1, 2]. However, DNN models face a challenge when they receive a continuous stream of information (Fig. 1a). One of the most critical issues in this context is catastrophic forgetting — a phenomenon during which the network suddenly loses the memory of previously learned items when learning new ones (Fig. 1b, top) [3-5]. This is due to the high plasticity of wirings in DNNs, which allows new inputs readily to overwrite old weight distributions. This raises a crucial issue known as the "stability-plasticity dilemma" [6], highlighting the need for a balance in DNNs between maintaining stability to preserve the memory of prior knowledge while simultaneously fostering flexibility to learn new inputs.

To resolve this issue, several studies developed revised models of DNNs that are resistant to catastrophic forgetting [7]. For example, some prior models suggested an algorithm in which new nodes or networks are added during the learning of new tasks or information [8-10]. Other models have incorporated memory replay or rehearsal strategies to enhance the resilience of DNN models against catastrophic forgetting [11-15]. However, these approaches not only require additional resources to store the old information but also accompany additional phases for the re-training of the old memory. Therefore, while these strategies partially address the problem of catastrophic forgetting, their complicated processing and heavy computational loads make them impractical. This emphasizes the necessity for new approaches that are more efficient and plausible.

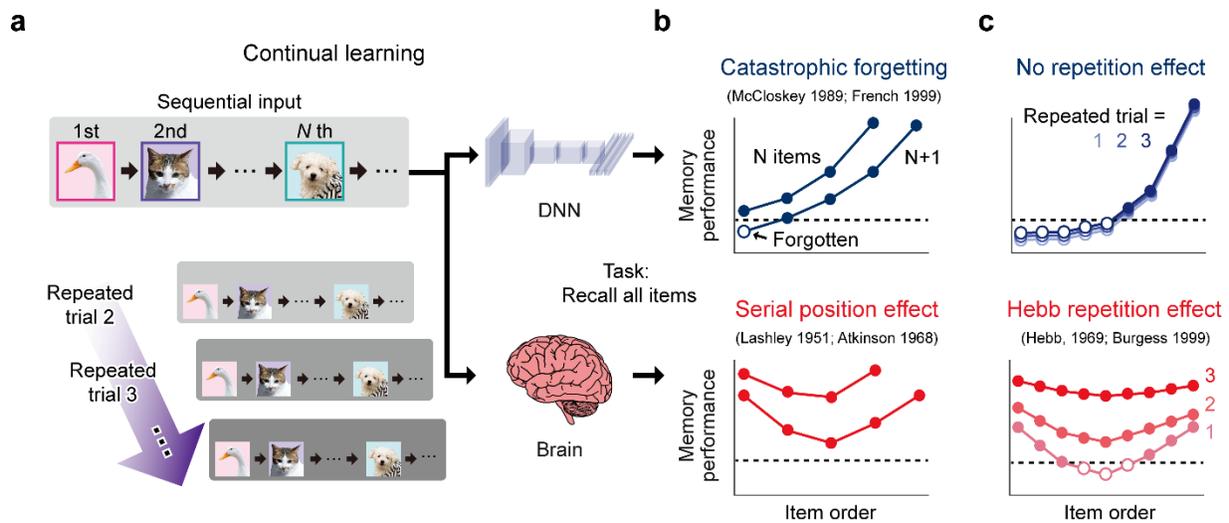

**Figure 1. Continual learning results from a conventional deep neural network and from the brain.** (a) Continual learning in DNNs and the brain. (b) Performance of each item after learning items continually. DNNs show high memorization performance only for a few recent items (catastrophic forgetting) while the brain retains its memory of both early and recent items (serial position effect). (c) Performance of each item after learning a sequence of items repetitively. The memory of DNNs does not improve upon repetition, whereas the brain enhances its memory of items after each repetition trial (Hebb repetition effect).

In contrast to the innate restrictions of DNNs in continual learning, human brains can balance stability and plasticity in their neural memory representations [16-18]. Unlike DNNs that overly weigh recent information, the brain adaptively allocates its memory resources to retain old inputs while reserving capacity for new information. A good signature of how the brain resolves the stability-plasticity dilemma is the "serial position effect" observed in sequential working memory tasks, in which items positioned at the beginning and end of a sequence are better memorized than those in the middle (Fig. 1b, bottom) [18-25]. Specifically, the combination of the "primacy effect" and "recency effect" enables a balanced weight between old and new information [26], achieving adaptive continual memory. Notably, this phenomenon persists when the total number of items in the sequence is varied [16, 27, 28]. When the total number of items increases, the memory resource for old items is reallocated to incorporate new ones, enabling the brain to learn new items while retaining old information. This illustrates the brain's ability to accomplish adaptive memory allocation even when the amount of information to be learned is uncertain. Another observation known as the "Hebb repetition effect" [29, 30] demonstrates that the brain can enhance a "weak" memory of a particular item by simply presenting it repeatedly (Fig. 1c, bottom). These observations may be important clues to understanding the brain's mechanism for adaptive, robust continual learning.

This leads to the question of how the brain implements the "serial position effect" and the "Hebb repetition effect," referring to the type of mechanism of synaptic plasticity that enables the network to retain both old and new information for adaptive continual learning. One possible candidate is synaptic metaplasticity [31-33], by which the plasticity of individual synapses are adjusted based on their modification history and current neural activity [34, 35]. In this way, the brain can stabilize certain synapses for old memory while letting others hold plasticity for new information, dealing with the stability-plasticity dilemma. A key advantage of this scenario is that no additional pre- or post-processing is necessary, unlike most previous models that still require computationally intensive procedures to determine the appropriate metaplastic states of each synapse [32]. Moreover, when the amount of information to learn is uncertain, such models are prone to forgetting previously encoded information, particularly when they learn new information that exceeds the network's memory capacity [31-33] — a limitation that the brain rarely faces.

Here, we propose a novel metaplasticity model by which the issues described above are fully addressed, i.e., where continual learning for sequential input is ensured in the complete absence of pre- or post-processing. Inspired by our previous notion of the distinct role of flexible and stable encodings in sequential working memory in the brain [16], our model synapses are designed to have a metaplasticity profile randomly sampled from a distribution ranging from extremely stable to extremely flexible values. This simple intermixing of distinct synapses could successfully replicate the key features of biological working memory for "catastrophic forgetting-free" continual learning in DNNs. We demonstrate that a network composed of both stable and unstable synapses enables DNNs to learn sequences of images, with high memory retention for both early and recent items, consistently maintaining sequential information of varying lengths. The model also exhibits an adaptive capacity-performance tradeoff and the frequency-dependent consolidation of repeated information, characteristics also observed in human working memory [19-21, 29, 30]. We demonstrate that this frequency-dependent consolidation enables robust memory against data poisoning attacks [36], distinguishing it from conventional DNNs.

Overall, our results show that a neuromimetic memory model inspired by human working memory characteristics enables catastrophic forgetting-free, continual learning in conventional DNNs, thus providing an effective model of adaptive continual learning without any pre- or post-processing.

## 2 Methods

### 2.1 Neural Network model

AlexNet [37] was used as a representative model of a deep neural network. This network consists of two parts: feature extraction (Input-Pool5) and classification (FC6-output) networks (Table S1). In detail, the feature extraction network consists of five convolutional layers and the classification network has three fully connected layers. The detailed parameters were sourced from Krizhevsky et al. (2012). To analyze the impact of our metaplastic rule on the retention of sequential task performance in fully connected layers, the feature extraction network (Input-Pool5) is pre-trained with the ImageNet dataset and frozen during the entire simulation [33]. In contrast, the classification network is randomly initialized from a Gaussian distribution with a zero mean and the standard deviation set to balance the input strength across the fully connected layers (bias = 0) [38].

### 2.2 Flexibility of the synapses

To realize stable and unstable synapses, we introduced the concept of "synaptic flexibility" to modulate the update of individual weights. Flexibility values range from 0 to 1, where a synapse with a flexibility of 0 is deemed fully stable, signifying that its weight remains unchanged. Conversely, a synapse with a flexibility of 1 is entirely unstable, allowing unrestricted updates without downscaled learning rates. This synapse operates identically to the synapses within conventional neural networks. The flexibility of the synapses was collectively set during the network's initialization before it was trained. The assigned flexibility of each synapse is retained throughout the training and testing phases. Specifically, we modulated the learning rate of a given weight, denoted as "$w$," by scaling it using the function $S$, which ranges in value from 0 to 1.

$$w_{updated} := w - [S(flexibility, \Delta w) \cdot \eta] \frac{\partial}{\partial w} J(w, b) \quad (1)$$

Here, $\eta$ is the learning rate and $J(w, b)$ is a loss function in terms of weight $w$ and bias $b$. $\Delta w$ represents the difference between the initial weight value and the weight value after the learning of the previous item. For instance, the $\Delta w$ value during the phase in which the network learns 3rd item is the difference between the initial weight and the weight immediately after learning 2nd item. $S$ is a function of the flexibility of the weight and $\Delta w$, and $\alpha$ is a hyperparameter which scales the width of $S$.

$$S(flexibility, \Delta w) = 1 - \tanh^2(\alpha \frac{1-flexibility}{flexibility} \cdot \Delta w) \quad (2)$$

### 2.3 Training Dataset

To facilitate the training of networks for the classification of an extended sequence of diverse image classes, we created a two-digit MNIST dataset [39]. This dataset was generated by combining handwritten images of single-digit numbers obtained from the MNIST dataset. The resulting two-digit MNIST dataset comprised data points wherein two different number images were aligned side by side, with the alignment centered. Each training dataset class encompassed 5,000 distinct images of the same number, while the corresponding test dataset classes were composed of 1,000 different images of the same number.

To manage interclass similarities, we deliberately omitted classes composed of a single digit, such as "00," "11," and "22" from both the training and test datasets. Additionally, when forming two-digit numbers, we ensured that the smaller number preceded the larger number, maintaining consistency. For instance, in the case of 0 and 1, "01" was included as a valid class, while "10" was excluded.

### 2.4 Continual Learning Task

To assess the continual learning capabilities of model networks, we devised a continual learning task inspired by the structure of sequential working memory tasks [18, 20, 21]. The continual learning task unfolds as a series of sequential tasks, where each task demands the network to identify images of a specific number in a sequential manner (Fig. 2c). For example, in the first task, the network is trained to classify images of the number 38 along with images of numbers that do not include 38 (referred to as non-38). Upon completing the first task, the network progresses sequentially to learn the next class of images, such as 89. This sequential learning process continues until the network goes through all classes within the designated learning sequence.

### 2.5 Test for memorized items versus forgotten items

To determine if a network memorizes a specific item in a sequence, we introduced the concept of a "memorized item". We examined whether a network shows significantly higher performance for each item than for items with randomly shuffled labels. Specifically, we generated one thousand different label-shuffled datasets by randomly permuting the labels of the original dataset and then estimated the network's performance on these shuffled datasets to obtain a thousand different "control" performance values. Then, we determined the "memorized item" by investigating whether the performance on a particular item is significant ($p<0.05$), i.e., whether the performance on the original dataset is higher than that on the shuffled datasets, at least for 950 out of 1,000 trials.

### 2.6 Memory Performance Measurement

After completing the learning of each class of various sample images, the network undergoes testing to classify all classes in the learning sequence. To evaluate the network's memory retention for each class, we measured "memorization performance," defined as the classification accuracy specific to each item. For example, the memorization performance on number 38 is the probability of the network correctly classifying an image of 38 as belonging to number 38. The significance of memory performance was tested by comparing performance values with the chance level, $1/N_{readouts}$.

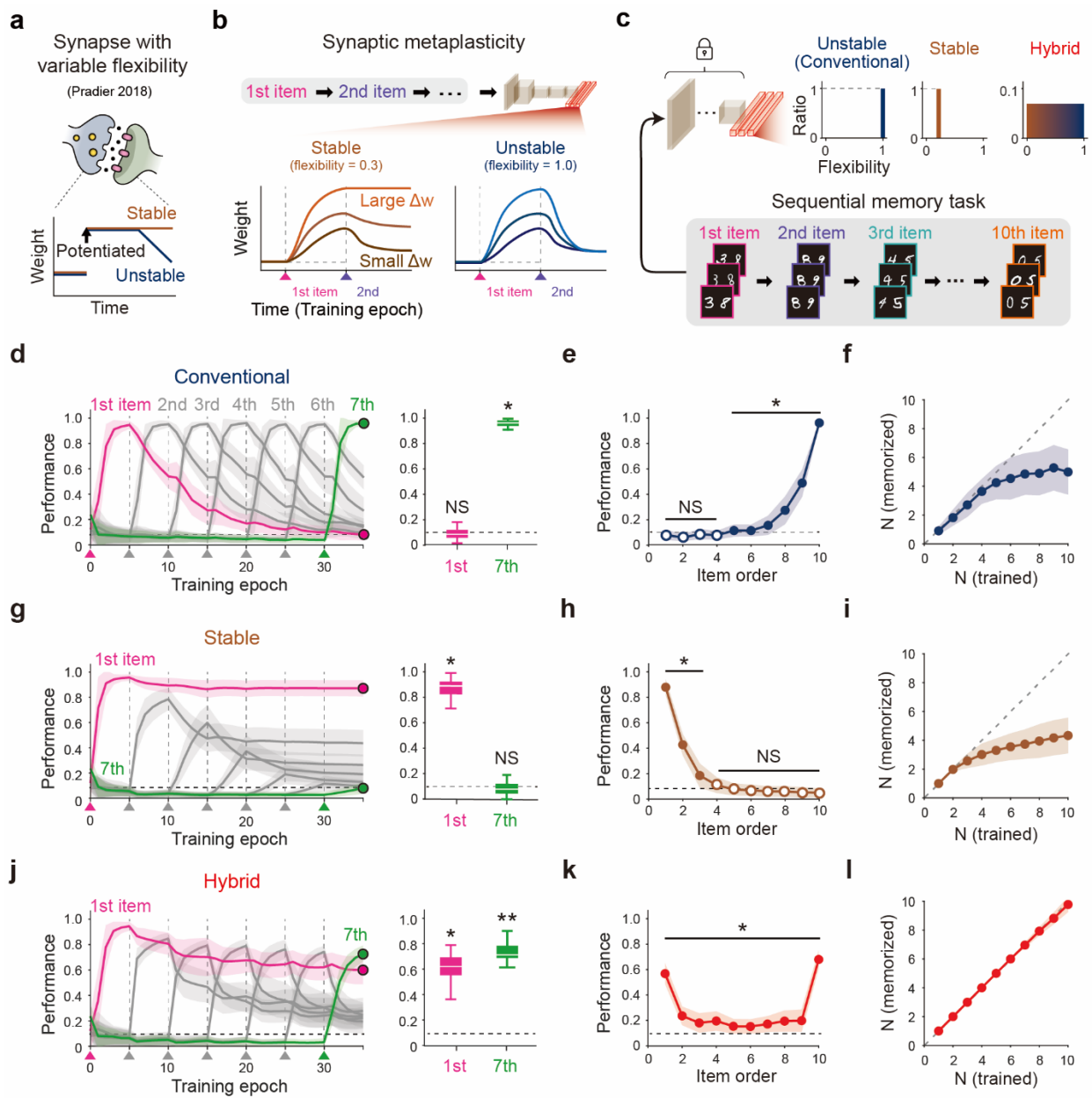

**Figure 2. The coexistence of stable and unstable synapses in the model network generates the serial position effect, as observed in human working memory.** (a) Labile LTP reported in mammalian brains. LTP can be either retained stably or be rapidly discharged. (b) Schematic diagram of weight updates of a neuromimetic synapse. Left: The weight of a stable synapse is fixed upon sufficient potentiation or depression. Right: The weight of an unstable synapse remains changeable during training. Pink and purple arrows indicate time points where 1st and 2nd items start to be trained, respectively. (c) Top: Model neural networks with different flexibility compositions; the conventional model is composed of unstable synapses for which flexibility = 1. The stable model is composed of stable synapses for which flexibility = 0.3. The hybrid model includes synapses with a uniform distribution of flexibility ranging from

0 to 1. Bottom: Design of a continual learning task; a sequence of items is sequentially trained on three models with different synaptic compositions, and after the entire training process is completed, the network is tested on its ability to classify the learned items. (d) Performance of the conventional network. Left: Change in the memory performance for the first seven items. Right: Performance on 1st and 7th items after learning 7th item (NS, p =1.00; *p = 1.67×10$^{-160}$, one-sided t-test, $N_{trial}$ = 100). (e) Memory performance of the conventional model after learning the entire item sequence. The conventional model shows the recency effect (NS, p = 1.00; *p < 1.00×10$^{-4}$, one-sided t-test, $N_{trial}$ = 100). Dashed lines denote the chance level. (f) Number of memorized items with respect to the total number of trained items. (g) Performance of the stable model. Left: Change in memory performance for the first seven items. Right: Performance on 1st and 7th items after learning 7th item (NS, p = 1.00; *p = 1.59×10$^{-109}$, one-sided t-test, $N_{trial}$ = 100). (h) Memory performance of the stable model after learning the entire item sequence. The stable model shows the primacy effect (NS, p > 0.08; *p < 2.23×10$^{-308}$, one-sided t-test, $N_{trial}$ = 100). (i) Number of memorized items with respect to the total number of trained items. (j) Performance of the hybrid network. Left: Change in memory performance for the first seven items. Right: Performance on 1st and 7th items after learning 7th item (*p = 1.65×10$^{-67}$; **p = 6.18×10$^{-101}$, one-sided t-test, $N_{trial}$ = 100). (k) Memory performance of the hybrid network after learning the entire item sequence. The hybrid model demonstrates the serial position effect, memorizing all items in the learning sequence (*p < 2.23×10$^{-308}$; one-sided t-test, $N_{trial}$ = 100). (l) Number of memorized items with respect to the total number of trained items. Data are presented as the mean±SD.

## 3  Results

### 3.1  Emergence of the serial position effect from the coexistence of stable and unstable synapses

To investigate the effects of incorporating both unstable and stable synapses on continual learning tasks, we employed a widely used deep neural network model, AlexNet [37]. We designed the metaplasticity of the synapses by adopting the characteristics of the labile long-term potentiation [40] (LTP) (Fig. 2a), a particular form of synaptic plasticity observed in the brain where synaptic weights potentiated by stimulation can be either maintained (stable) or discharged (unstable) depending on the conditions. Accordingly, we defined the concept of "synaptic flexibility" (see Methods for details, see also Supplementary Fig. 1); for a stable synapse, the weight can change initially but is retained stably when largely deviating from the initial value during continual learning (Fig. 2b, left). On the other hand, an unstable synapse is kept flexible, allowing its weight to change continuously throughout the training process (Fig. 2b, right). Based on our earlier works [16, 17], we hypothesized that stable synapses enable the retention of previous memories while unstable synapses facilitate the learning of new information. We thus expected that stable synapses would contribute to the generation of the primacy effect (i.e., items presented first in the sequence are memorized better) while unstable synapses would facilitate the recency effect (i.e., items presented last are memorized better); a model with layers that are intermixed with stable and unstable synapses, therefore, would show the serial position effect [16].

To test the role of each type of synapse in continual learning, we introduced the model synapse into the fully connected layers of AlexNet, devising three distinct networks with varying synaptic compositions (Fig. 2c, top). First, we established an unstable network, where the flexibility of weights within fully connected layers was all set to the maximum value, 1, as in conventional deep neural networks. Subsequently, we designed a stable network, where the flexibility of synaptic nodes was very low (fixed at 0.3 in the current model) but still allowed learning. Finally, we modeled a hybrid network in which synapses with various degrees of flexibility coexist. The flexibility value of each synapse is randomly sampled from a uniform distribution ranging from 0 to 1. Among those possible designs to implement this condition, we mimicked the configuration observed in the brain, where synapses with varying degrees of plasticity coexist within a single area [41]. To assess each model's capability to retain memory across both new and previous inputs, we designed a continual learning task (Fig. 2c, bottom) in which the model network was sequentially trained to memorize and classify ten distinct handwritten numbers [39] (see Methods for details). For instance, in the first task, the network was trained to distinguish between handwritten images of 1st item (the number 38) and non-1st item (another number). Subsequently, the network progressed to learn another two-digit number (2nd item) among a set of diverse two-digit numbers. This learning process was repeated to train ten different numbers sequentially.

After training on the 1st item, all three models demonstrated high performance precision for classifying 1st item. However, for the following items, the responses of the models diverged notably. After learning 7th item, the unstable network (conventional DNN) exhibited high classification performance for 7th item but completely lost its ability to classify 1st item (Fig. 2d; 1st item vs. chance level, NS, $p = 1.00$; 7th item vs. chance level, *$p = 1.67 \times 10^{-160}$, one-sided t-test, $N_{trial} = 100$). When the learning all ten items was complete, the conventional model showed the recency effect, retaining the last six items in the learning sequence to a degree significantly higher than the chance level (Fig. 2e; NS, $p = 1.00$; *$p < 1.00 \times 10^{-4}$, one-sided t-test, $N_{trial} = 100$). As a result, the network memorized only up to six out of ten items, indicating typical catastrophic forgetting (Fig. 2f; see Methods for details). On the other hand, the stable network managed to retain its proficiency for old items but failed to learn new items. For example, after training 7th item, it still maintained the performance of 1st item above the chance level but started failing to achieve optimal performance for the last item (Fig. 2g; 1st item vs. chance level, *$p = 1.59 \times 10^{-109}$; 7th item vs. chance level, NS, $p = 1.00$, one-sided t-test, $N_{trial} = 100$). After learning ten items, the proficiency of the network remained high for the first three items but decreased for the subsequent items (Fig. 2h; *$p < 2.23 \times 10^{-308}$; NS, $p > 0.08$, one-sided t-test, $N_{trial} = 100$). As a result, it memorized approximately four of the oldest items only, exhibiting the primacy effect (Fig. 2i). These results show that the conventional and the stable models fail to store sequential items when the number of items exceeds a certain threshold.

In contrast, we found that the hybrid model successfully balanced learning and retention, achieving high classification performance for all of the items in the sequence — learning novel items while retaining the memory of older items (Fig. 2j; 1st item vs. chance level, *$p = 1.65 \times 10^{-67}$; 7th item vs. chance level, **$p = 6.18 \times 10^{-101}$, one-sided t-test, $N_{trial} = 100$). The network demonstrated the serial position effect, which enables it to maintain its performance level for all ten items above the chance level (Fig. 2k; *$p < 2.23 \times 10^{-308}$, one-sided t-test, $N_{trial} = 100$) such that the number of memorized items is identical to the total number of items (Fig. 2l). The hybrid model demonstrated strong capability in retaining the memory of both new and old information, thus showing the potential to endow neural networks with the two conflicting capabilities of learning and memorizing. It is notable that the serial position effect can emerge consistently with various hybrid network designs, i.e., different profiles of the combination between stable and unstable synapses (Supplementary Fig. 2). For example, a network in which all synapses have a single, intermediate value of flexibility (e.g., 0.8), referred to here as the "single-value hybrid network model," also exhibits a serial position effect. In this case, the profile of the serial position effect can readily be manipulated by a slight change of this single value, suggesting possible applications in which the memorization characteristics of the network must be adaptively controlled (Supplementary Fig. 2). However, in general, the simplest means of implementing a hybrid network is to sample the flexibility value of each synapse randomly from a uniform distribution of all possible values; such a model also allows the convenient manipulation of the serial position effect profile by biasing the distribution of the flexibility values towards either the stable or unstable side, resulting in a corresponding primacy-heavy or recency-heavy effect (Supplementary Fig. 3).

### 3.2 The serial position effect is robustly preserved even when the item number varies

Next, to examine the robustness of the hybrid model's serial position effect when the amount of information varies, we sequentially trained the model with 30 readout nodes to learn 30 different items (Fig. 3a). We found that the hybrid model maintained the serial position effect regardless of the length of the sequence, adeptly retaining both old and new information (Fig. 3b, top). Notably, the model sustained memory even for items that exhibited the lowest performance (in the middle of the sequence) well above the chance level, while also successfully learning an unknown number of novel items. In contrast, the conventional model showed catastrophic forgetting, failing to memorize early items in the sequence (Fig. 3b, bottom). As a result, the hybrid model memorized significantly more items than the conventional model (Fig. 3c; *p = 3.70×10$^{-63}$, two-sample t-test, $N_{trial}$ = 100). This result demonstrates the hybrid model's dynamic utilization of memory resources, flexibly allocating a portion of resources to retain old items while using the remainder to encode novel information without any supervision or control algorithm. This caused the average performance across the items to decrease (Fig. 3d) but simultaneously allowed the model to encompass more items above the chance level. This adaptive capability of the hybrid model was further confirmed through estimation of the gross memory (Fig. 3e), calculated by summing the memory performance values of all items included in the sequence. We found that the gross memory increased gradually in both models as the number of trained items increased, but it appeared significantly higher in the hybrid model than in the conventional model (*p = 2.03×10$^{-57}$, two-sample t-test, $N_{trial}$ = 100). Notably, the gross memory in the hybrid model asymptotically converged to a constant value when the number of trained items exceeded a certain value. However, the network could still accommodate more items in its memory by decreasing the average performance of individual items, demonstrating the capacity-performance tradeoff (Fig. 3e). This result highlights the model's adaptability in realistic conditions, where the amount of information to be learned is uncertain.

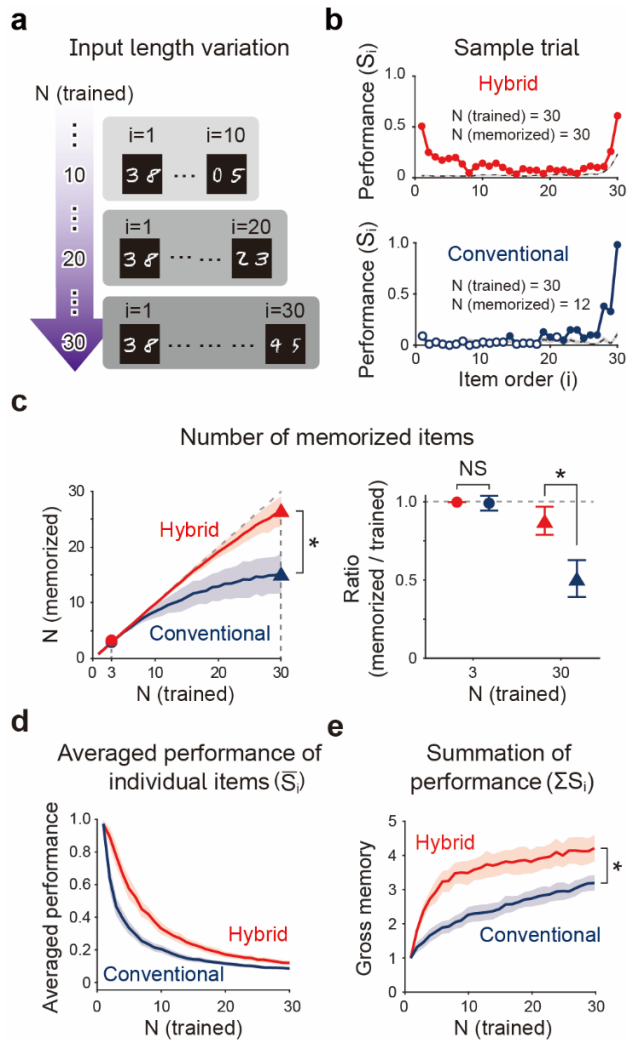

**Figure 3. The hybrid network robustly exhibits the serial position effect as the length of the learning sequence increases.** (a) The network model learns various learning sequence lengths. (b) A sample memory performance curve after learning sequences composed of 30 items. Top: The serial position effect is retained in the hybrid model, memorizing all the items in the sequence. Bottom: The conventional network showed catastrophic forgetting. Solid circles denote memorized items, while open circles indicate forgotten items (See Methods). (c) Left: Number of memorized items of the hybrid and conventional models for various learning sequence lengths. The number of memorized items of the conventional model converges as the length of the input sequence increases, representing a fixed capacity of the network (NS, $p = 0.16$; *$p = 3.70 \times 10^{-63}$, two-sample t-test, $N_{trial} = 100$). Right: The ratio of memorized items to the total number of items. (d) Average memory performance on entire items in the hybrid and the conventional models $(\overline{S_i})$. $S_i$ denotes the performance of $i$ th item in the sequence. (e) Summation of the individual item's performances in hybrid and conventional models $(\sum_{i=1}^{N(total)} S_i)$. The gross memory, the summation of the memory performance values of all items included in the sequence, is significantly higher in the hybrid model than in the conventional model (*$p = 2.03 \times 10^{-57}$; two-sample t-test, $N_{trial} = 100$). Data are presented as the mean±SD.

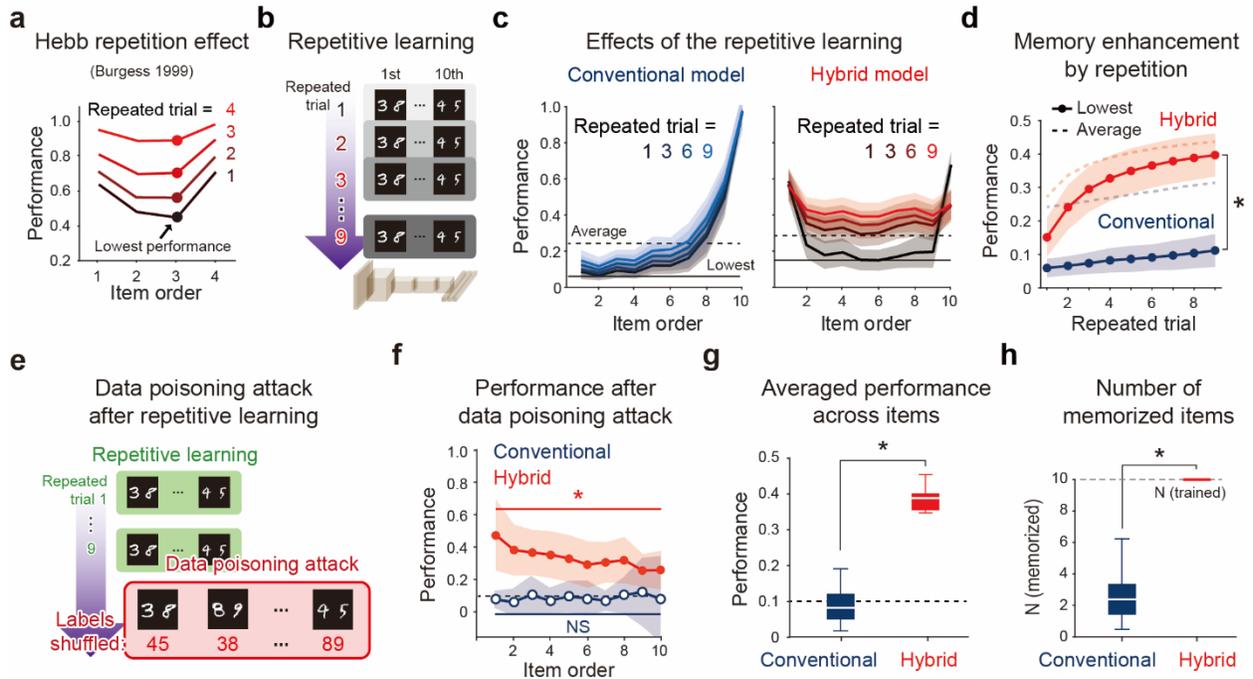

**Figure 4. Sequential memory performance can be improved via repetitive learning in the hybrid network.** (a) Repetitive learning can improve the memory of the sequence in the brain (Hebb repetition effect). (b) Repetitive learning task used to train the model neural network. (c) Performance curve after repetitive learning. Left: The performance of the conventional model does not significantly change after repetitive learning. Right: The performance of the hybrid model is improved upon repetitive learning. Dashed lines denote the average performance value across the items after learning the first repeated trial, and solid lines show the lowest performance value after learning of the first repeated trial. (d) Performance on an item with the lowest level of performance in the learning sequence (*$p = 7.32 \times 10^{-91}$, two-sample t-test, $N_{trial} = 100$). (e) Simulation scheme with erroneous data included as the last learning repetition trial. Networks are trained with items with correct labels for the first nine repeated trials, followed by shuffled label data learning upon the tenth repeated trial. (f) Performance curve of the hybrid and conventional models after the learning of erroneous data (Conventional, NS, $p > 0.14$; Hybrid, *$p < 2.23 \times 10^{-308}$, one-sided t-test, $N_{trial} = 100$). (g) Averaged performance across items within the sequence after learning the erroneous data (*$p = 7.52 \times 10^{-48}$, two-sample t-test, $N_{trial} = 100$). (h) Number of the memorized items after learning of erroneous data from the conventional and the hybrid models (*$p = 2.32 \times 10^{-38}$, two-sample t-test, $N_{trial} = 100$). Data are presented as the mean±SD.

### 3.3 Memory performance is improved with repeated training

In the hybrid model, memory performance appears mostly higher for items at the sequence's beginning and end, whereas those positioned in the middle have relatively lower performance outcomes. This results in some items in the middle of the sequence being forgotten, as the total number of items within the sequence increases significantly.

Accordingly, we looked into whether a learning mechanism described from brain research can improve this memory performance profile of the hybrid model. "Hebb repetition learning" [29, 30] is the idea that repeated exposure to information strengthens the memory of things within a sequence and reduces forgetfulness (Fig. 4a). With this notion, we examined if the performance can be improved by repeated training in our model network. We trained the hybrid model in nine repetitive trials, using an identical sequence each time, and analyzed the memory performance of each item within the learning sequence (Fig. 4b). We found that the performance of the conventional DNN does not depend on the repetition of training (Fig. 4c, left). In contrast, in the hybrid model, repetitive review led to a substantial enhancement in memory, particularly for items in the middle of the sequence, as observed during Hebb repetition learning (Fig. 4c, right). From the analysis of the lowest performance outcome within the sequence after each repeated trial, we found that the hybrid network exhibited a more significant improvement in the lowest performance across repeated learning, compared to the conventional model (Fig. 4d; *$p = 7.32 \times 10^{-91}$, two-sample t-test, $N_{trial}$ = 100). This led to a narrowed difference between the lowest and highest performance values in the sequence, facilitating the alignment of memory performances across various items (Supplementary Fig. 4). These findings suggest that the coexistence of stable and unstable synapses equips neural networks with the ability to improve memory through repeated training, achieving leveled performance outcomes for all items regardless of their position in the sequence. This "repetition-based memory enhancement" emphasizes the model's capability to undertake dynamic resource allocation and implies the ability to utilize previously learned information to facilitate ongoing learning — a crucial aspect of continual learning.

Inspired by this result, we anticipated that our model could effectively filter out unreliable information based on how frequently the information appears. By assuming that frequently presented information is likely to be relatively reliable while less frequent items may be more susceptible to errors [42], we hypothesized that the hybrid model could robustly maintain its task performance level even after a data poisoning attack [36] once it is trained correctly with repeated data. To test this hypothesis, we trained two models with nine consecutive repetitions of correctly labeled data, followed by a single training sequence in which the data labels were shuffled [43, 44] (Fig. 4e). We found that the conventional model lost the memory of all ten items after learning the incorrectly labeled data (Fig. 4f, blue; NS, $p > 0.14$, one-sided t-test, $N_{trial}$ = 100). In contrast, the hybrid model maintained the memory of all ten items after the data poisoning attack (Fig. 4f, red; *$p < 2.23 \times 10^{-308}$, one-sided t-test, $N_{trial}$ = 100), showing a higher average performance level after the data poisoning attack, compared to the conventional model (Fig 4g; *$p = 7.52 \times 10^{-48}$, two-sample t-test, $N_{trial}$ = 100). Furthermore, the number of memorized items after the data poisoning attack showed that the hybrid model retained significantly more items compared to the conventional model (Fig. 4h; *$p = 2.32 \times 10^{-38}$, two-sample t-test, $N_{trial}$ = 100). Overall, these results suggest that the intermixing of stable and unstable synapses in the neural network enables the automatic filtration of information, allowing the network to discern and prioritize data based on its frequency or reliability.

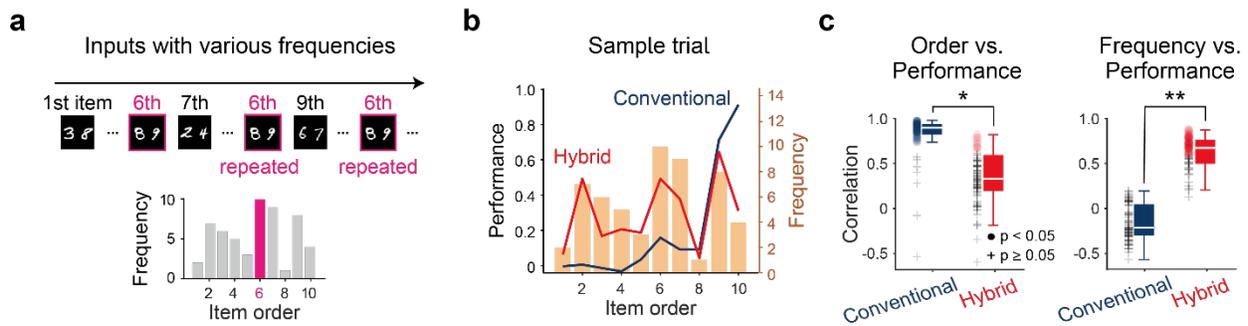

**Figure 5. The hybrid network adaptively allocates memory resources based on the input frequency.** (a) Simulation scheme while varying the input frequency; the training frequencies of each item within the sequence differ. (b) Example of memorization performance for an item with respect to the corresponding order within the sequence and repetition frequency. The performance of the conventional model does not depend on the learning frequency, whereas the performance of the hybrid model depends on the learning frequency (Conventional, r(frequency) = 0.25, p = 0.49; Conventional, *r(order) = 0.82, p = 3.40×10$^{-4}$; Hybrid, *r(frequency) = 0.92, p = 1.67×10$^{-4}$; Hybrid, r(order) = 0.10, p = 0.79, Spearman correlation). (c) Left: Spearman correlation between an item's order and its memorized performance. The conventional network showed a higher correlation than the hybrid network (*p = 2.29×10$^{-21}$, two-sample t-test, $N_{trial}$ = 100). Right: Spearman correlation between the item's repetition frequency and its memorized performance. The hybrid network showed a high correlation, while none of the trials from the conventional network showed a significant performance-frequency correlation (**p = 2.51×10$^{-56}$, two-sample t-test, $N_{trial}$ = 100).

### 3.4 Memory performance is selectively enhanced for frequently presented items

Our findings demonstrate the hybrid model's proficiency in memory enhancement through repeated training, which could prompt further explorations into potential applications of the model. Specifically, here we hypothesized that the model could selectively enhance the memory of frequently presented items while sacrificing the memory of less frequent items. To test this idea, the hybrid model underwent training with ten distinct items, each featuring varied presentation frequencies ranging from 1 to 10 (Fig. 5a). We expected that the hybrid model would exhibit enhanced memorization for frequently presented items compared to less frequent items, whereas the conventional model would undergo catastrophic forgetting and shows the item-order dependent memory performance found earlier.

As expected, the performance of the conventional model exhibited a strong positive correlation with the item order, whereas the hybrid network's performance tended to show a higher correlation with the item's presentation frequency for various network initialization conditions and for the combinations of items composing the learning sequence (Figs. 5b and c). We found that the correlation between the item's performance and its order was significantly higher in the conventional model than in the hybrid model (Fig. 5c, left; Conventional vs. Hybrid, *p = 2.29×10$^{-21}$, two-sample t-test, $N_{trial}$ = 100). In contrast, the correlation between performance and presentation frequency appeared significant in the hybrid model, but none of the trials from the conventional network showed a significant correlation (Fig. 5c, right; Conventional vs. Hybrid, **p = 2.51×10$^{-56}$, two-sample t-test, $N_{trial}$ = 100). Overall, the memory performance of the conventional DNN depends heavily on each item's position within the sequence rather than its presentation frequency, whereas the hybrid network allocates its resources to more frequent items, relying less on their position in the sequence. This result suggests that the hybrid model automatically "forgets" less-used information, allowing the freed memory resources to enhance the memory of more frequently appearing, probably more important information.

## 4 Discussion and Conclusion

We showed that the random mixing of synapses with varying degrees of flexibility could give rise to brain-like characteristics in continuous learning, such as the serial position effect and the Hebb repetition effect. We discovered that DNNs with such connections are able to replicate the serial position effect observed in the brain by maintaining memory for both recent and old items. As a result, the network could continuously learn sequential inputs without catastrophic forgetting, even when the input length varies unexpectedly. Furthermore, we discovered that the network could naturally bring about the frequency-dependent consolidation of repeated information and the adaptive capacity-performance tradeoff, which enables the network adaptively to use memory resources for sequential input and further be robust against data poisoning attacks. Lastly, we showed that the performance level of individual items within the sequence can also be manipulated by controlling their training frequencies.

Our findings carry significance as they suggest the viability of a straightforward approach — simply intermixing stable and unstable synapses without modulating the physical structure of conventional model networks — to replicate the aspects of continual learning observed in the brain, providing a solution to the problem of catastrophic forgetting in DNNs. Our model carries significant convenience in that it can be implemented in pre-existing neural networks, compared to prior algorithms. Specifically, our model does not necessitate external memory storage beyond the primary network used for continual learning to store the information of previously learned items [15, 45-48]. Furthermore, it avoids alterations to the network architecture, such as additions of extra nodes to accommodate new information [8, 9]. Additionally, it imposes less of a computational burden, such as the calculation of a Fisher information matrix for each task or the tracking of the history of weight changes, to determine which synapses should be stabilized and which should be released [31, 32]. This simplicity and flexibility in implementation allow us to integrate our model in accordance with any specific requirements and constraints.

Our model specifically possesses features that are advantageous for real-world applications. First, the model achieves the "capacity-performance tradeoff" without requiring additional training of information or structural modifications of the network, as noted above (Fig. 3). In sequential learning, the model dynamically reallocates resources from previously learned items to newly acquired ones to retain information pertaining to both items. This feature is particularly useful when the number of items is uncertain and/or the significance of each item within the sequence is unknown. Under such conditions, conventional networks often prioritize the last few items, potentially overlooking previous items and thus struggling to accommodate new information or retain old information as the model encounters extensive data exceeding its capacity. In contrast, our model voluntarily modifies the accuracy level of each item to accommodate a greater number of items as needed, balancing capacity and performance. Second, our model is able selectively to filter out potentially erroneous data, particularly beneficial when working with datasets containing incorrect information or noise (Fig. 4). Drawing from the principle that frequently presented information is often more reliable [42], our model selectively strengthens the memory of frequently encountered data while attenuating the influence of less common, potentially erroneous data, while conventional networks mechanically prioritize recently learned data. This feature of our model enhances the network's robustness and reliability when managing datasets with different levels of quality and significance. Lastly, our model not only memorizes information, but also adaptively "forgets" less frequent information (Fig. 5). Our model reallocates memory resources from "rarely learned" information to "frequently learned" information, consequently reinforcing the memory of frequently presented data. This finding implies that the memory performance curve of the hybrid model can be manipulated by the user based on their needs. Specifically, users can purposefully train certain information they want the model to memorize strongly at higher frequencies while training less important information at lower frequencies. This adaptability is lacking in the conventional network that prioritizes recent information regardless of its training frequency. Thus, the features of our model provide substantial benefits in real data applications.

Our model mirrors the structural characteristics observed in the cortical regions of the brain, where varying degrees of synaptic flexibility coexist. This type of organization may be beneficial for robust memory functions [49, 50], particularly in dynamically changing environments. Our simulations of conventional and stable networks highlight the distinct functions of unstable synapses and stable synapses (Fig. 2), showing that unstable synapses primarily contribute to learning and memorizing recent items, whereas stable synapses mainly retain the memory of early items. These findings also align with experimental observations of the caudate nucleus [41], further validating the biological relevance of our model's synaptic mechanisms. Previous studies of stable and unstable encodings in the caudate nucleus reported that flexible coding in the caudate head distinctively represents unstable (or recent) information, while stable (or older) experiences are encoded in the caudate tail [41, 51]. Moreover, it was reported that the caudate body encompasses synapses with both stable and flexible coding patterns [41], similar to our model of the hybrid network. These findings imply that a random distribution of synaptic flexibilities, without specific regulation or complex calculations, can readily result in brain-like continual learning. This model provides an effective and plausible scenario for organizing neural circuits for continual learning given that such conditions may be achievable spontaneously during the brain's developmental stages. Furthermore, our model offers a plausible explanation of how the brain resolves the stability-plasticity dilemma and accomplishes robust continual learning. With a simple synaptic rule, our model proposes a feasible neural mechanism underlying cognitive phenomena such as the serial position effect and the Hebb repetition effect, which are hallmarks of sequential working memory. While a number of previous models face limitations in that they often require certain artificial manipulation strategies, such as modulating the neuronal gain in a specific manner or heavy computations for pre- or post-processes [52, 53], our model demonstrates that a simple synaptic-level organization—a mix of stable and unstable synapses—successfully reproduces particular key mechanisms of continual learning.

Our findings imply that the brain-like ability of continual learning can be achieved from the coexistence of stable and unstable synapses, but further investigations and additional evidence are needed to fully understand this scenario. One notable difference between our model and the brain is the absence of recurrent connections among nodes, a factor considered crucial for working memory retention in biological systems. Given that our study utilized fully connected feedforward networks to train image classification tasks, this may not precisely mirror the complexities of biological neural networks. However, the findings from our model also suggest that neuronal layers, each wired with feedforward connections with randomized flexibilities, are sufficient for generating a sequential memory profile similar to that observed in human working memory. Possibly, extra recurrent wirings may add temporal fine tuning of neuronal activities to model the neurological phenomena observed in the human working memory system [54], though this can be further examined in subsequent studies. Notably, our results leverage the strengths of computational model simulations, allowing for extensive testing with large datasets [55-58], repeated iterations, and an analysis of circuit components at a single-unit level [59-61], tasks which are virtually impossible in experimental studies. Our simulations enable the manipulation of relevant circuit structures to test key hypotheses, including the control of learning rates for different weights based on their flexibility. Despite the acknowledged disparities from biological systems, our computational model provides valuable insights, demonstrating that brain-like continual learning features can manifest in neural networks incorporating both stable and unstable encodings.

## Acknowledgments

This work was supported by a grant from the National Research Foundation of Korea (NRF) funded by the Korean government (MSIT) (No. NRF-2022R1A2C3008991 to S.P.).

# Supplementary Information

**Table S1. Summary of the AlexNet architecture used in the present study**

| Layer | Type | Number of units | Kernels | Activations |
|---|---|---|---|---|
| **Input** | Image input | $227 \times 227 \times 3$ | - | - |
| **Conv1** | Convolution | $55 \times 55 \times 96$ | Weights $11 \times 11 \times 3 \times 96$ Bias $1 \times 1 \times 96$ | ReLU and cross-channel normalization |
| **Pool1** | Max pooling | $27 \times 27 \times 96$ | - | - |
| **Conv2** | Convolution | $27 \times 27 \times 256$ | Weights $5 \times 5 \times 48 \times 256$ Bias $1 \times 1 \times 256$ | ReLU and cross-channel normalization |
| **Pool2** | Max pooling | $13 \times 13 \times 256$ | - | - |
| **Conv3** | Convolution | $13 \times 13 \times 384$ | Weights $3 \times 3 \times 256 \times 384$ Bias $1 \times 1 \times 384$ | ReLU |
| **Conv4** | Convolution | $13 \times 13 \times 384$ | Weights $3 \times 3 \times 192 \times 384$ Bias $1 \times 1 \times 384$ | ReLU |
| **Conv5** | Convolution | $13 \times 13 \times 256$ | Weights $3 \times 3 \times 192 \times 256$ Bias $1 \times 1 \times 256$ | ReLU |
| **Pool5** | Max pooling | $6 \times 6 \times 256$ | - | - |
| **FC6** | Fully connected | $1 \times 1 \times 4096$ | Weights $4096 \times 9216$ Bias $4096 \times 1$ | ReLU and dropout |
| **FC7** | Fully connected | $1 \times 1 \times 4096$ | Weights $4096 \times 4096$ Bias $4096 \times 1$ | ReLU and dropout |
| **FC8** | Fully connected | $1 \times 1 \times N_{readouts}$ | Weights $N_{readouts} \times 4096$ Bias $N_{readouts} \times 1$ | Softmax |
| **Output** | Classification output | - | - | - |

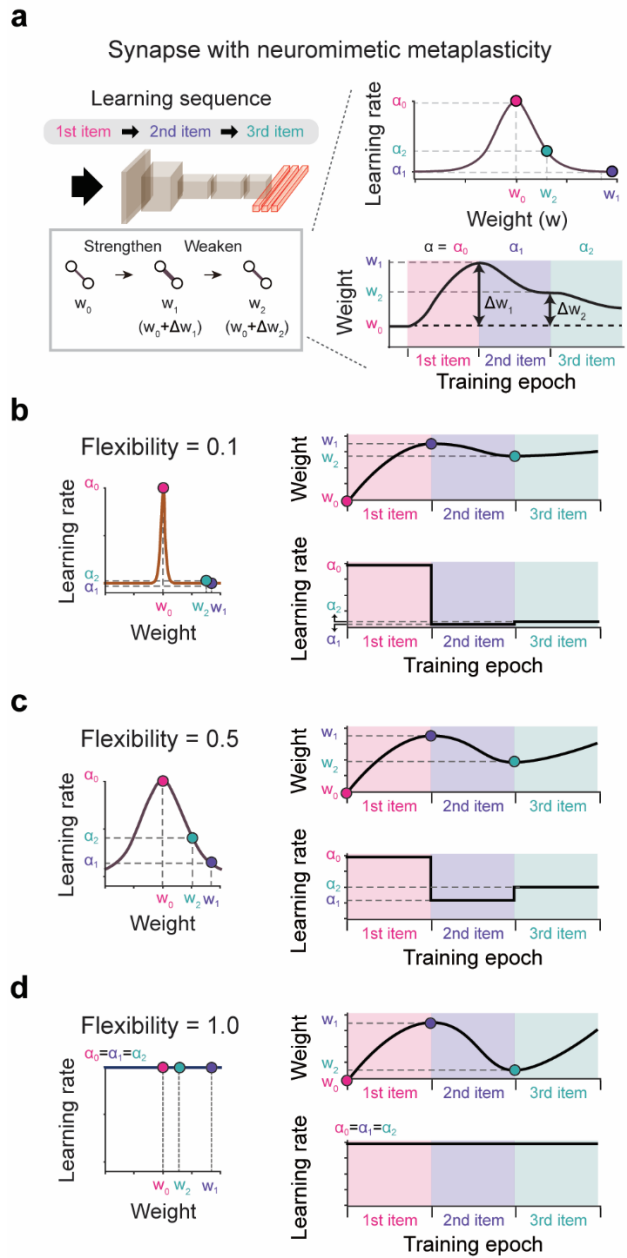

**Figure S1. Weight change dynamics by synaptic flexibility.** (a) Schematics of the weight update rule. (b-d) Learning rate functions of synapses with different levels of flexibility (left), and corresponding weight and learning rate change curves during sequential learning (right). Sample weight changes of synapses with flexibility 0.1 (b), 0.5 (c), and 1.0 (d) are shown.

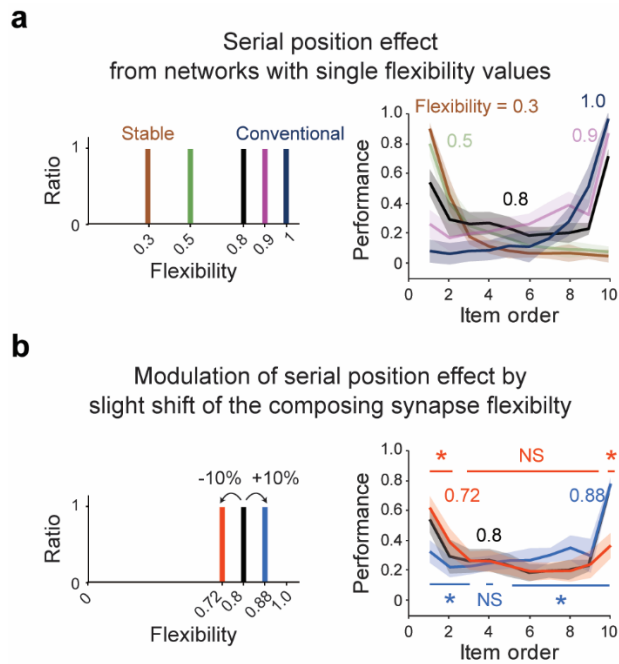

**Figure S2. Profiles of the serial position effect resulting from various unified values of synaptic flexibility.** (a) Serial position effect curves resulting from networks with unified flexibility values set to 0.3, 0.5, 0.8, 0.9, and 1.0. In this model, flexibility values assigned to synapses are identical across all synapses in a network. (b) Serial position effect curves for which the flexibility values are 0.72, 0.8, and 0.88. The profiles of the serial position effect curves are significantly influenced by small changes in the flexibility value in the network models with a single flexibility value (for flexibility = 0.72, NS, $p > 0.34$; *$p < 2.23 \times 10^{-308}$; for flexibility = 0.88, NS, $p = 0.10$; *$p < 0.01$, one-sided t-test, $N_{trial} = 50$). Data are presented as the mean$\pm$SD.

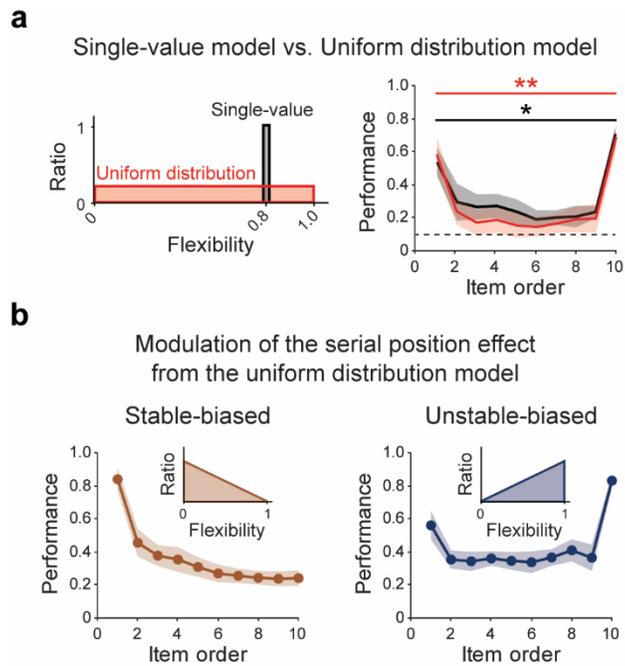

**Figure S3. Profiles of the serial position effect resulting from a uniform distribution of synaptic flexibility.** (a) Serial position effect curves of a unified model with a single value of flexibility set to 0.8 and a model with distributed flexibility, i.e., where the flexibility is uniformly distributed between 0 and 1. Both networks commonly exhibit balanced serial position effect curves and memorize all ten items in the sequence. (Model performance vs. chance level, *p < 0.01; **p < $2.23 \times 10^{-308}$, one-sided t-test, $N_{trial}$ = 50). The dashed line denotes the chance level. (b) Modulation of the serial position effect resulting from biasing the flexibility distribution of the uniform distribution model. Left: Serial position effect curves resulting from networks with flexibility distributions biased towards the stable side, derived from the hybrid model. The degree of bias positively correlates with the strength of the primacy effect observed. Right: Serial position effect curves resulting from networks with flexibility distributions biased towards the unstable side, derived from the hybrid model. The degree of bias positively correlates with the strength of the recency effect observed. Data are presented as the mean±SD.

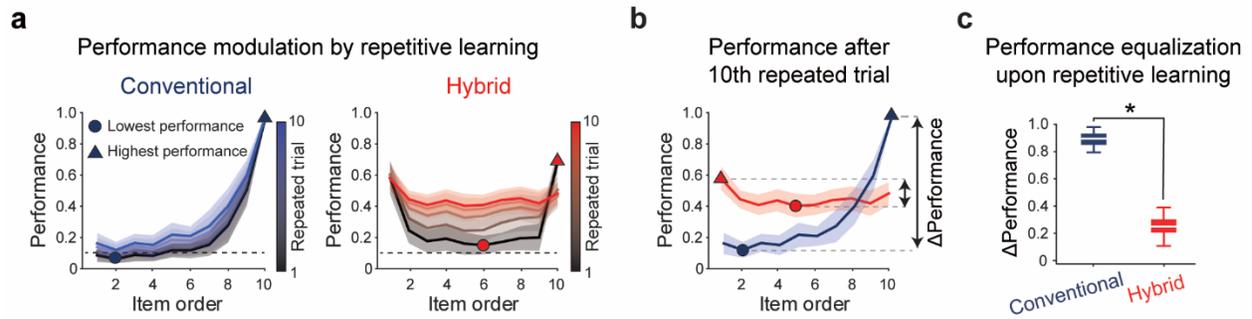

**Figure S4. Improvement of the minimum performance within a sequence resulting from repetitive learning.** (a) Serial position effects resulting from repetitive learning. Left: Performance from the conventional model. Right: Performance from the hybrid model. (b) Memory performance of the hybrid and conventional models after learning ten repetitive trials. (c) The discrepancy between the highest and lowest performance values (ΔPerformance) after ten repetitive trials. The ΔPerformance from the hybrid model is significantly lower than that from the conventional model, demonstrating that repetitive learning equalizes memory performances of individual items within the sequence (*p =2.65×10$^{-144}$, two-sample t-test, N$_{trial}$ = 100).